\documentclass[twocolumn,10pt,a4paper]{article}

\usepackage{natbib}
\usepackage{booktabs} 

\usepackage[utf8]{inputenc}

\usepackage{hyperref}

\usepackage{booktabs}
\usepackage[colorinlistoftodos]{todonotes}
\usepackage{tikz}
\usetikzlibrary{positioning, calc}

\graphicspath{{figs/}}

\usepackage{xspace}
\newcommand{\eg}{e.g.,\xspace}
\newcommand{\etal}{et al.\xspace}
\newcommand{\ie}{i.e.,\xspace}

\title{The Free-play Sandbox: a Methodology for the Evaluation of Social
Robotics and a Dataset of Social Interactions}

\author{Séverin Lemaignan, Charlotte Edmunds, Emmanuel Senft, Tony Belpaeme \\
Plymouth University \\
Plymouth, United Kingdom \\
Email: \tt firstname.lastname@plymouth.ac.uk
}

\begin{document}

\maketitle

\abstract{

Evaluating human-robot social interactions in a rigorous manner is notoriously
difficult: studies are either conducted in labs
with constrained protocols to allow for robust measurements and a degree of
replicability, but at the cost of ecological validity; or \emph{in the wild},
which leads to superior experimental realism, but often with limited
replicability and at the expense of rigorous interaction metrics.

We introduce a novel interaction paradigm, designed to elicit rich
and varied social interactions while having desirable scientific properties
(replicability, clear metrics, possibility of either autonomous or Wizard-of-Oz
robot behaviours). This paradigm focuses on child-robot interactions, and
builds on a sandboxed free-play environment. We present the
rationale and design of the interaction paradigm, its 
methodological and technical aspects (including the open-source
implementation of the software platform), as well as two large open datasets
acquired with this paradigm, and meant to act as experimental baselines for
future research.}

\begin{figure}
    \centering
    \includegraphics[width=0.9\columnwidth]{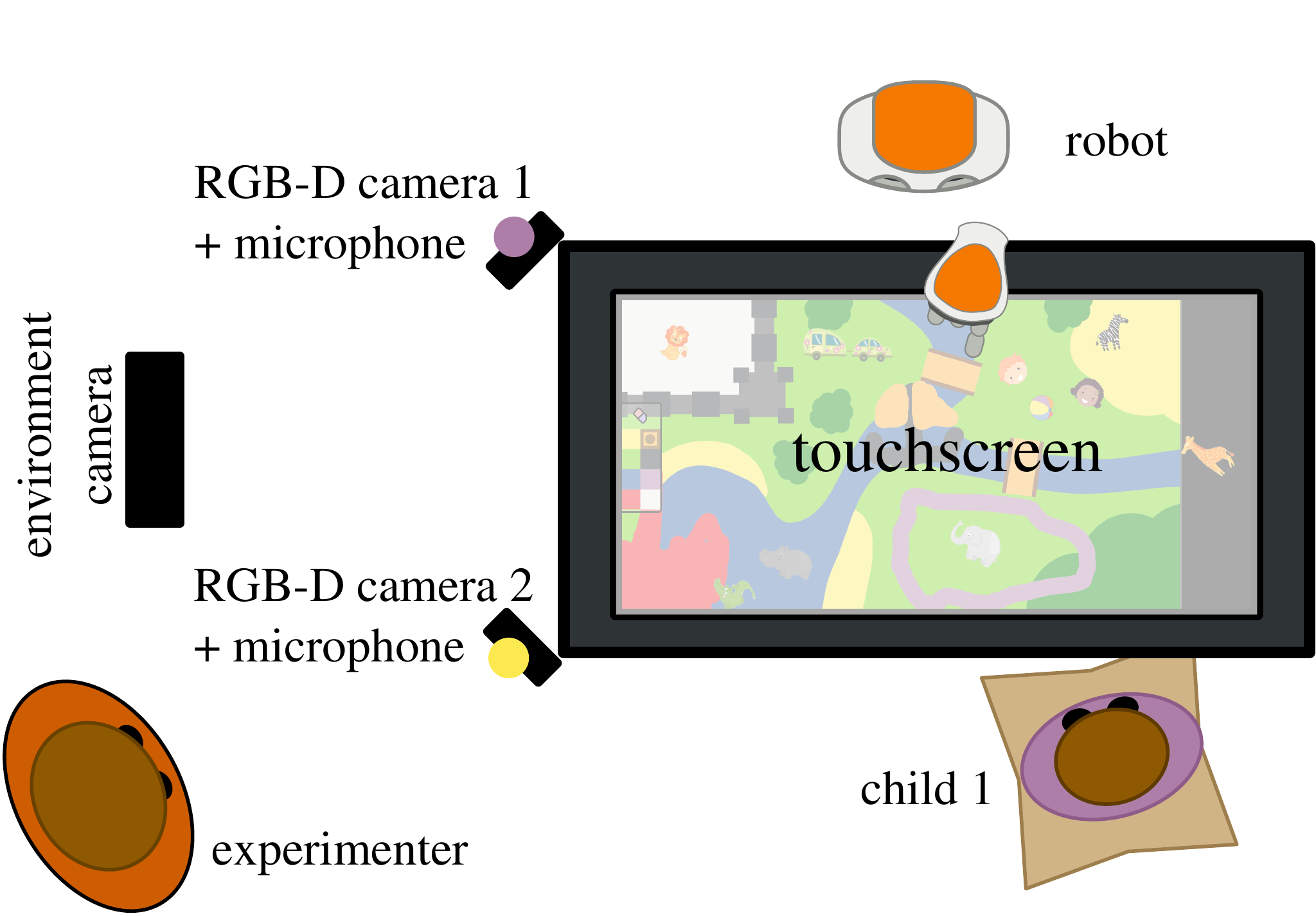}
    \caption{The free-play social interactions sandbox: two children interact in
    a free-play situation, by drawing and manipulating items on a touchscreen.
    Children are facing each other and sit on cushions. Each child wears a
    bright sports bib, either purple or yellow, to facilitate later
    identification.}

    \label{fig|freeplay}
\end{figure}

\section{The challenges in evaluating social interactions}
\label{sec:intro}

\subsection{Studying social interactions}

Studying social interactions requires a social \emph{situation} that
effectively elicits interactions between the participants. Such a situation is
typically scaffolded by a social task, and consequently, the nature of this task
influences in fundamental ways the kind of interactions that might be observed
and analysed. In particular, the socio-cognitive tasks commonly found in the
literature of experimental psychology (and HRI) often have a narrow focus:
because they aim at studying one (or a few) specific social or cognitive skills
in isolation and in a controlled manner, these tasks are typically simple and
highly constrained (for instance, an object hand-over task; a perspective-taking
task with cubes, etc.). While these focused endeavours are important and
necessary, we -- as a community -- also acknowledge that these interaction
scenarios do not reflect the complexity and dynamics of real-world
interactions~\cite{baxter2016characterising}, and we certainly observe a strong
trend within our community towards capturing, interpreting and acting upon the rich set of
naturally-occurring social interactions.

Specifically, we believe that further progress in the study of human-robot
interactions should be scaffolded by socio-cognitive challenges that:

\begin{itemize}
    \item are long enough and varied enough to elicit a large range of
        interaction situations;
    \item foster rich multi-modal interaction, such as simultaneous speech, gesture, and gaze
        behaviours;
    \item are loosely directed, to maximise natural, non-contrived behaviours;
    \item evidence complex social dynamics, such as rhythmic coupling, joint attention,
        implicit turn-taking;
    \item include a certain level of non-determinism and unpredictability.
\end{itemize}

The challenge lies in designing a social task that exhibits these features
\emph{while maintaining `good' scientific properties} (repeatability, replicability,
robust metrics) as well as good practical properties (not requiring unique or
otherwise very costly experimental environments, not requiring very specific
hardware or robotic platform, easy deployment, short enough experimental
sessions to allow for large groups of participants).

In this paper, we introduce such a task, designed to elicit rich, complex, varied
social interactions while being well suited for interactions with robots and
supporting rigorous scientific methodologies.

\subsection{Social play}

Our interaction paradigm is based on free and playful interactions (free play)
in a \emph{sandboxed} environment: while the interaction is free (participants
are not directed to perform any particular task beyond playing), the activity is
both \emph{scaffolded} and \emph{constrained} by the setup mediating the
interaction (essentially, a large table-top touchscreen).
Participant engage in open-ended and non-directive play situations, yet sufficiently
well defined to be reproducible and practical to record and analyse.

This initial description frames the socio-cognitive interactions that might be
observed and studied: playful, dyadic, face-to-face interactions. While gestures
and manipulations (including joint manipulations) play an important role in this
paradigm, the participants do not typically move much during the interaction.
Because it builds on play, this paradigm is also naturally suited to the study
of child-child and child-robot interactions.

The choice of a playful interaction is supported by the wealth of social
situations and social behaviours that \emph{play} elicits. Most of the research
in this field builds on the early work of Parten who established five
\emph{stages of play}~\cite{parten1932social}, corresponding to different stages
of development, and accordingly associated with typical age ranges:

\begin{enumerate}
    \item {\bf Solitary (independent) play}, age 2-3: Playing separately from
        others, with no reference to what others are doing.
    \item {\bf Onlooker play}, age 2.5-3.5: Watching others play. May engage in
        conversation but not engage in doing. True focus on the children at
        play.
    \item {\bf Parallel play} (adjacent play, social co-action), age 2.5-3.5: Playing
        with similar objects, clearly beside others but not with them (near
        but not with others.)
    \item {\bf Associative play}, age 3-4:  Playing with others without
        organization of play activity. Initiating or responding to
        interaction with peers. 
    \item {\bf Cooperative play}, age 4+: Coordinating one's behavior with that
        of a peer. Everyone has a role, with the emergence of a sense of
        belonging to a group. Beginning of "team work."
\end{enumerate}

These five stages of play have been extensively discussed and refined over the
last century, yet remain remarkably widely accepted as such. It must be noted
that the age ranges are only indicative. In particular, most of the early
behaviours still occur at times by older children.

Interestingly, these five stages can been looked at from the perspective of HRI
as well. They certainly evoke a roadmap for the development of human-robot
social interactions.


\section{The Free-play Sandbox}
\label{sec:freeplay}

\subsection{Task}

We have designed a new experimental task, called the \emph{free-play sandbox},
that is based on free play interactions. Pairs of children (4-8 years old) are
invited to freely draw and interact with items displayed on an interactive
table, without any explicit goal set by the experimenter
(Fig.~\ref{fig|freeplay}).  The task is designed so that children can engage in
open-ended and non-directive play, yet it is sufficiently constrained to be
suitable for recording, and allows the reproduction of social behaviour by an
artificial agent in comparable conditions.

\begin{figure}[ht!]
    \centering
    \includegraphics[width=0.9\columnwidth]{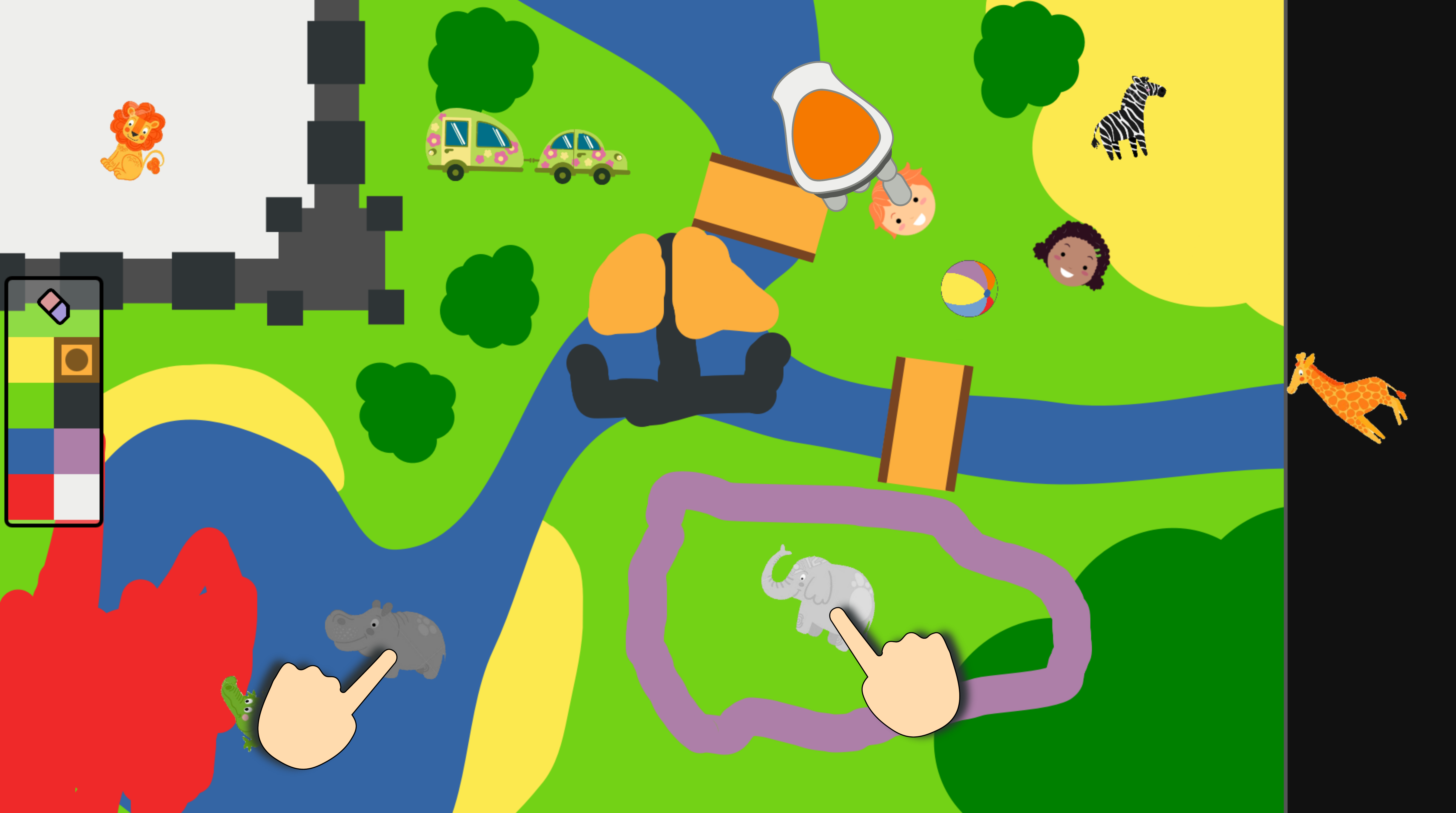}
    \caption{Example of a possible game situation. Items (animals,
    characters...) can be dragged over the whole play area, while the background
    picture can be painted over by picking a colour.}

    \label{fig|sandbox}
\end{figure}

The free-play sandbox follows the sandtray
paradigm~\cite{baxter2012touchscreen}: a large touchscreen (60cm $\times$ 33cm,
with multitouch support) is used as an interactive surface (\emph{sandtray}).
Two children play together by freely moving interactive items on the surface
(Fig.~\ref{fig|sandbox}). A background image depicts a generic empty
environment, with different symbolic colours (water, grass, beach, bushes...).
By drawing on top of the background picture, the children can change the
environment to their liking. The players do not have any particular task to
complete, they are simply invited to freely play. Importantly, they can play for
as long as they wish (for practical reasons, we have limited the sessions to a
maximum of 40 minutes in our own experiments, see Section~\ref{sec:dataset}).

Capturing all the interactions taking place during the play sessions is possible
and practical with this setup. Even though the children will typically move a
little, the task is fundamentally a face-to-face, spatially delimited,
interaction, and as such simplifies the data collection. For instance, during
our dataset acquisition campaign (120 children, more than 45h of footage), the
children's faces were automatically detected in 98\% of the recorded frames (see
Section~\ref{sec:dataset}).

\subsection{Applications}
\label{sec:applications}

\paragraph{Child-Child Interaction}

The free-play sandbox provides the opportunity to observe children interacting
in a natural way in an open but framed setup. As the system can run on a single
computer platform it can easily be deployed in the 'wild', in places where the
children naturally interact such as classroom. The quantity and thoughtfulness
of information logged allows to keep a track of every interaction happening
around the game. 

These advantages combined with the openness of the task proposed make this setup
a powerful tool to observe and quantify a large spectrum of social behaviours
expressed by children when interacting in a natural environment (might be
interesting to add a list here). The compactness of the system makes it easy to
compare data from different locations.

\paragraph{Child-Robot Interaction}
\label{ssec|CRI}

This free-play sandbox provides the opportunity to explore child-robot
interactions in this open, real world environment as shown in Figure
\ref{fig|freeplay}. 

Depending of the focus of the study, two modes of control for the robot are
available. If the interest is on evaluating a specific robot behaviour, the
robot can be autonomously controlled using inputs from the different sensors.
This setup allows to explore the impact of different social behaviours on the
children independently of the `game policy' controlling by the robot. 

On the other hand, if the focus is on the child behaviour and the technical
aspect is of a lower importance, the robot can be controlled by a human rather
than an algorithm. This paradigm, where the robot is tele-operated to interact
with a naive partner is called Wizard of Oz (WoZ) and is used in numerous
studies to explore the psychologic side of HRI \cite{riek2012wizard}. 

\paragraph{Deep Learning}

With the quantity of data logged and the high number of interaction achievable
with the free-play sandbox, it supports the type of requirement for recent
Machine Learning approaches such as deep learning. The similar position of the
children in all interactions makes the combination of data from different
interaction easier than other less compact systems.

From the information collected on the children, social behaviours can be
extracted and used on a robot.



\section{Implementation}
\label{sec:impl}

The software-side of the free-play sandbox is entirely
open-source\footnote{Source code: \url{https://github.com/freeplay-sandbox/core}}. It is
implemented using two main frameworks: Qt
QML\footnote{\url{http://doc.qt.io/qt-5/qmlapplications.html}} for the graphical
interface of the game, and the \emph{Robot Operating System} (ROS) for the
modular implementation of the data processing and behaviour generation
pipelines. The graphical interface interacts with the decisional pipeline over a
bidirectional QML-ROS bridge that we have developed for that purpose.

\begin{figure*}
    \centering
    \includegraphics[width=\linewidth]{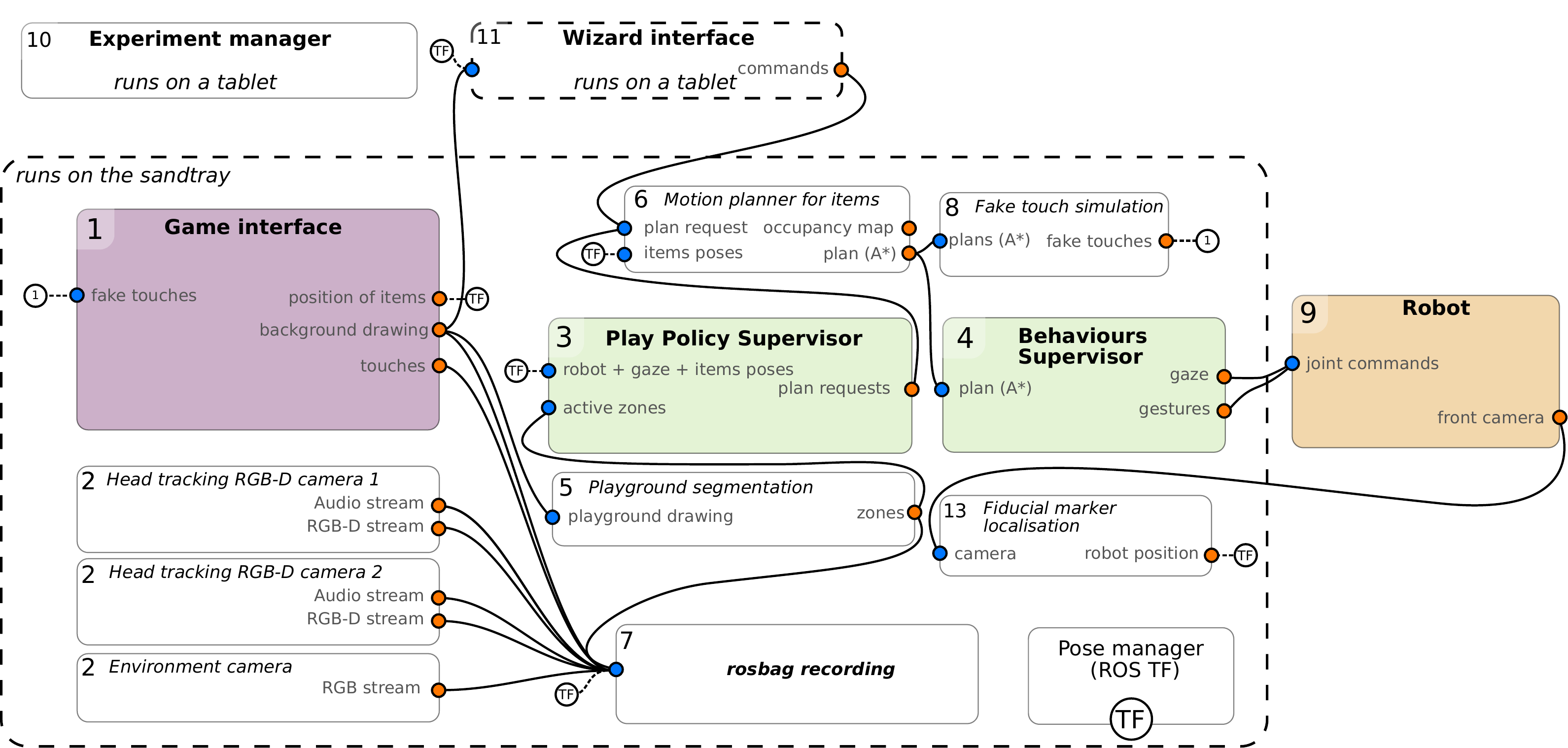}
    \caption{Software architecture of the free-play sandbox. Left (purple) nodes
    are connected to the sandtray (game interface (1) and camera drivers (2)). Nodes in the
    centre (green) implement the behaviour of the robot (play policy (3) and robot
    behaviours (4)). Several helper nodes are available, in particular, segmentation of the
    children drawings into zones (5), A* motion planning for the robot to move
    in-game items (6). Nodes are implemented in
    Python (except for the game interface, developed in QML) and inter-process
    communication relies on ROS. 6D poses are managed and exchanged via ROS TF.}

    \label{fig|architecture}
\end{figure*}

Figure~\ref{fig|architecture} presents the software architecture of the sandbox.

\begin{figure}
    \centering
    \includegraphics[width=\linewidth]{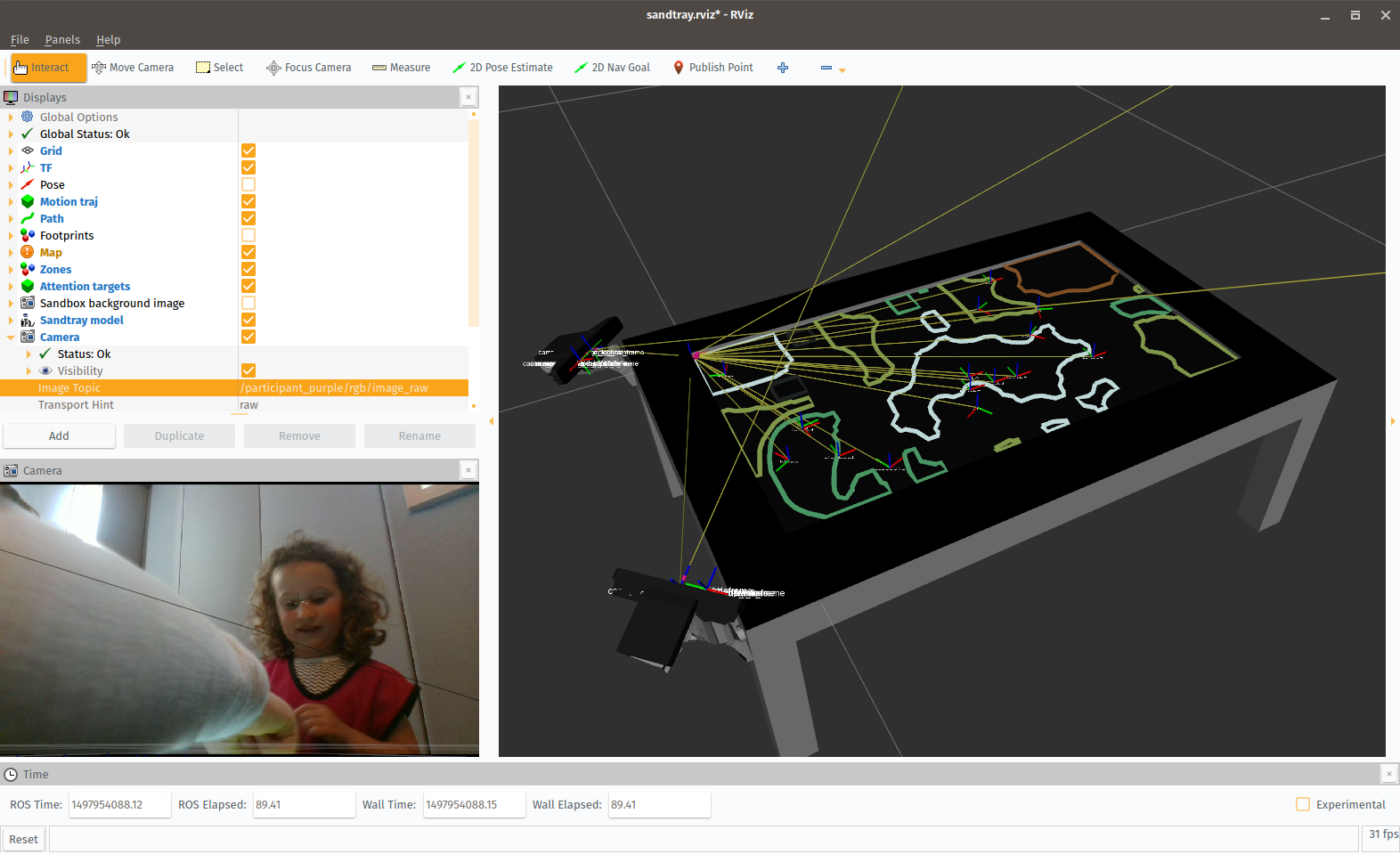}
    \caption{The free-play sandbox, viewed at runtime within ROS RViz. Simple
    computer vision is used to segment the background drawings into zones
    (visible on the right panel). The poses and bounding boxes of the
    interactive items are published as well, and turned into an occupancy map,
    used to plan the robot's arm motion.} 

    \label{fig|rviz}
\end{figure}

\subsection{Interactive game} The interactive game
(Fig.~\ref{fig|architecture}.1) is coded using QML, and displays a main
background image on top of which items (animals, humans and objects) can be
moved. The children can also use a drawing mode to create coloured strokes on a
layer between the background and the items, which adds another layer of
unconstrained interaction to the game (Figure~\ref{fig|sandbox}). The game
exposes the image of the background, the drawings, and the positions of the
objects as ROS TF frames.

\subsection{Sensing} Two Intel RealSense SR300 RGB-D cameras are mounted at
fixed positions on the sandtray frame, with custom designed 3D-printed brackets
that ensure that the cameras are oriented towards the children's face. Because
the cameras are rigidly mounted onto the sandtray's frame, their accurate
geometric transformations with respect to the sandtray screen are known.
Combined with hardware calibration, it allows for accurate localisation of the
children and in particular, children's faces. In addition to the images, both
cameras can perform stereo audio recording. One ROS node per camera
(Fig.~\ref{fig|architecture}.2) publishes on dedicated topics the audio and video streams.

A third `external' (and non-calibrated) camera is usually used as well to record
the environment of the experiment with a wider angle (\emph{environment camera}
in Figure~\ref{fig|freeplay}).

\subsection{Robot Control} As stated in section~\ref{ssec|CRI}, a robot
(Fig.~\ref{fig|architecture}.9) can act as play partner instead of one of the
children. This robot can either be autonomous selecting actions based on the
inputs provided by the sensors and the game or be controlled by a human in a
Wizard of Oz fashion.

\paragraph{Autonomous} The current implementation exposes a large number of
information on the game and the state of the child that can be used in the robot
controller. The position of every item is exposed as a TF frame, the background
is segmented in zones of identical colors (Fig.~\ref{fig|architecture}.5),
social element of the state the interaction are collected through the RGBD
camera and the microphone facing the child. As visible on
Figures~\ref{fig|freeplay} and~\ref{fig|rviz}, the camera covers the head of the
child as well as most of the upperbody, and applying libraries such as DLib and
OpenPose, the position of facial feature and skeleton of the child are extracted
and can be used to obtain: head gaze, gaze and gestures such as pointing. All
these inputs can be combined to provide the robot with more social inputs to
test the sociability of a robotic controller (Fig.~\ref{fig|architecture}.3) and
its impact on the interaction.

The robot's location is obtained by displaying fiducial markers on the
touchscreen before the start of the interaction, so the transformation between
the robot coordinate system and the touchscreen is known
(Fig.~\ref{fig|architecture}.13). And this robot location can also be used to
identify gazes from the child to the robot. 

To make the children believe the robot is moving objects on the touchscreen, we
synchronise a moving pointing gesture of the robot
(Fig.~\ref{fig|architecture}.4) and a series of fake touches
(Fig.~\ref{fig|architecture}.8) appied on the screen, moving the desired object.
Once an object and a goal position have been selected, a planner
(Fig.~\ref{fig|architecture}.6) generate a path for this image using the
A\mbox{*} algorithm on an occupancy map obtained with the items footprints, then
this plan is sent to a nodes synchronising the actuation on the robot and the
fake touches on the game.

Other actions such as gaze, pointing or speech are also exposed as simple ROS
topics.

\paragraph{Wizard-of-Oz} To allow an experimenter to control the robot, a GUI to
control the robot (Fig.~\ref{fig|architecture}.11) is provided and presents an
identical representation of the state of the game on an other application which
can be used on a tablet for example. The wizard can drag the objects in a
similar fashion as what the child would do on the Sandtray, and on the release,
the robot executes the dragging motion on the Sandtray, moving an object to a
new location. The source code can be easily modified to add new specific buttons to
execute other actions, such as having the robot talk to the child.


\subsection{Experiment Manager}

We have developed as well a dedicated, web-based, interface can be used by the experimenter
to manage the whole experiment and data acquisition procedure
(Fig.~\ref{fig|architecture}.10). This interface ensures that all the required
software nodes are running, allow the experimenter to check the status and, if
needed, to start/stop/restart any of them. It also help managing large data
collection campaigns by providing a convenient web interface (usually used by
the experimenter on a tablet) to record the demographics, resetting the game
interface after each session, and automatically enforcing the acquisition
protocol (see Table~\ref{tab|protocol}).

This interface has been extensively used to acquire the dataset that we present at
Section~\ref{sec:dataset}.

\section{Canonical procedures for data collection \& analysis}

The section presents \emph{canonical} procedures to acquire data during testing,
to pre-process it, and analyse it. We call them \emph{canonical} because they
are standard procedures, and where relevant, well integrated into the software pipeline of the sandbox (\eg ROS
integration) and represent state-of-the-art techniques. For the specific purpose
of manually annotating the social interaction, we introduce as well a novel
coding scheme, resulting from the synthesis of several existing techniques
(Section~\ref{sec|coding-scheme} below).

However, these procedure are not normative. Researchers interested in
reusing the free-play sandbox task for their own research would naturally adapt
and extend these protocols to their own needs. Besides, certain aspects (most
notably, the audio processing) are yet to be properly investigated.

\subsection{Protocol}

We typically adhere to the acquisition procedure described in
Table~\ref{tab|protocol} with all participants.  To ease later identification,
each child is also given a different and brightly coloured sports bib to wear.

Importantly, during the \emph{Greetings} stage, we show the robot both moving and
speaking (for instance, ``Hello, I'm Nao. Today I'll be playing with you.
Exciting!'' while waving at the children). This is meant to set the children's
expectations: they have seen that the robot can speak, move, and even behave in
a social way.

Also, the game interface of the free-play sandbox offers a tutorial mode, used
to ensure the children know how to manipulate items on a touchscreen and draw.
In our experience, this has never been an issue for children.

\newcommand{\tabitem}{~~\llap{\textbullet}~~}

\begin{table}[!]
\caption{Data acquisition protocol}
    \label{tab|protocol}
\centering
\begin{tabular}{p{\linewidth}}
\toprule
\bf Greetings \emph{(about 5 min)} \\
\tabitem explain the purpose of the study: showing robots how children play  \\
\tabitem briefly present a Nao robot: the robot stands up, gives a short
message, and sits down. \\
\tabitem place children on cushions \\ 
\tabitem complete demographics on the tablet \\
\tabitem remind the children that they can withdraw at anytime \\ \midrule
\bf Tutorial \emph{(1-2 min)}  \\
explain how to interact with the game, ensure the children are confident
with the manipulation/drawing \\ \midrule
\bf Free-play task (up to 40 min)\\
\tabitem initial prompt: \emph{"Just to remind you, you can use the animals or draw. Whatever you
like. If you run out of ideas, there's also an ideas box. For example, the first one is a
zoo. You could draw a zoo or tell a story. When you get bored or don't want to play
anymore, just let me know."} \\
\tabitem let children play \\
\tabitem once they wish to stop, stop recording \\ \midrule
\bf Debriefing      \emph{(about 2 min)} \\
\tabitem answer possible questions from the children \\
\tabitem give small reward (\eg stickers) as a thank you \\ \bottomrule
\end{tabular}
\end{table}

\subsection{Data collection}

Table~\ref{table|datastreams} lists the datastreams that are collected during
the game. By relying on ROS for the data acquisition (and in particular the
\texttt{rosbag} tool), we ensure all the $\approx$10 streams are synchronised,
timestamped, and, where appropriate, come with calibration information (for the
cameras mainly). In our experiments, cameras were configured to stream in qHD
resolution (960$\times$540 pixels) in an attempt to balance high enough
resolution with tractable file size. It results in \emph{bag} files weighting
$\approx$1GB per minute.

In our own experiments, all the data (including up to 5 simultaneous video
streams) was recorded on a single computer (quad core i7-3770T, 8GB RAM)
equipped with a fast 4TB SSD drive. This computer was also running the game
interface on its touch-enabled screen (sandtray), making the whole system
compact and easy to deploy (one single device).

\begin{table}[]
\centering
\caption{List of datastreams typically recorded. Each datastream is timestamped
with a synchronised clock to facilitate later analysis.}
\label{table|datastreams}
\begin{tabular}{@{}ll@{}}
\toprule
\bf Domain  & \bf Type                               \\ \midrule
children    & audio                                  \\
            & face (RGB + depth)                     \\
robot       & full 3D pose                           \\
environment & RGB                                    \\
touchscreen & background drawing (RGB)               \\
            & touches                                \\
            & position and orientation of in-game items         \\
 \multicolumn{2}{l}{static transforms between touchscreen and facial cameras} \\
 \multicolumn{2}{l}{cameras calibration informations}                         \\ \bottomrule
\end{tabular}
\end{table}

\subsection{Data processing}

\paragraph{Face and body pose analysis}

Off-line post-processing can be done on the images obtained from the cameras. We
rely on the CMU OpenPose library~\cite{cao2017realtime} to extract for both
children the upper-body skeleton, 70 facial landmarks including the pupil
position, as well as the hands' skeleton (when visible).

Further processing is possible: As the position of the
camera, a potential robot and any object on the game is known, this landmarks
can be mapped to high level behaviours such as pointing or looking at an object.
Additional analysis can be done on the facial landmarks to other social states,
such as main emotion felt by the child.

\paragraph{Audio processing}

Similar processing can be applied on the audio stream. Library such as OpenSMILE
provide audio features such as pitch and loudness contour, which inform on the
general state of the child.

As of today, no reliable speech recognition engine exists for
children~\cite{kennedy2017child}, but in the future, the audio should provide
textual information on the requests and comments produced by the child.

\paragraph{Game interactions analysis}

Game features are also produced by the different nodes involved in the analysis
of the game. The Playground segmentation produce a map of the regions based on
the colour which can be used with the positions of the animal to identify from
which zone to which zone an animal has been moved. The relative position of
animal can also indicate if two animals have been moved closer. These relations
and the drawing inform on what high level action the child is doing and can be
used to infer the child's goal or desire.

\subsection{Annotation of Social interactions}
\label{sec|coding-scheme}

Annotating social interaction beyond surface behaviours is generally difficult.
The observable, surface behaviours typically result of a superposition of the
complex and non-observable underlying cognitive and emotional states. As
such, these deeper socio-cognitive states can only be indirectly observed,
and their labelling is typically error prone.

\begin{figure}
    \centering
    \includegraphics[width=\columnwidth]{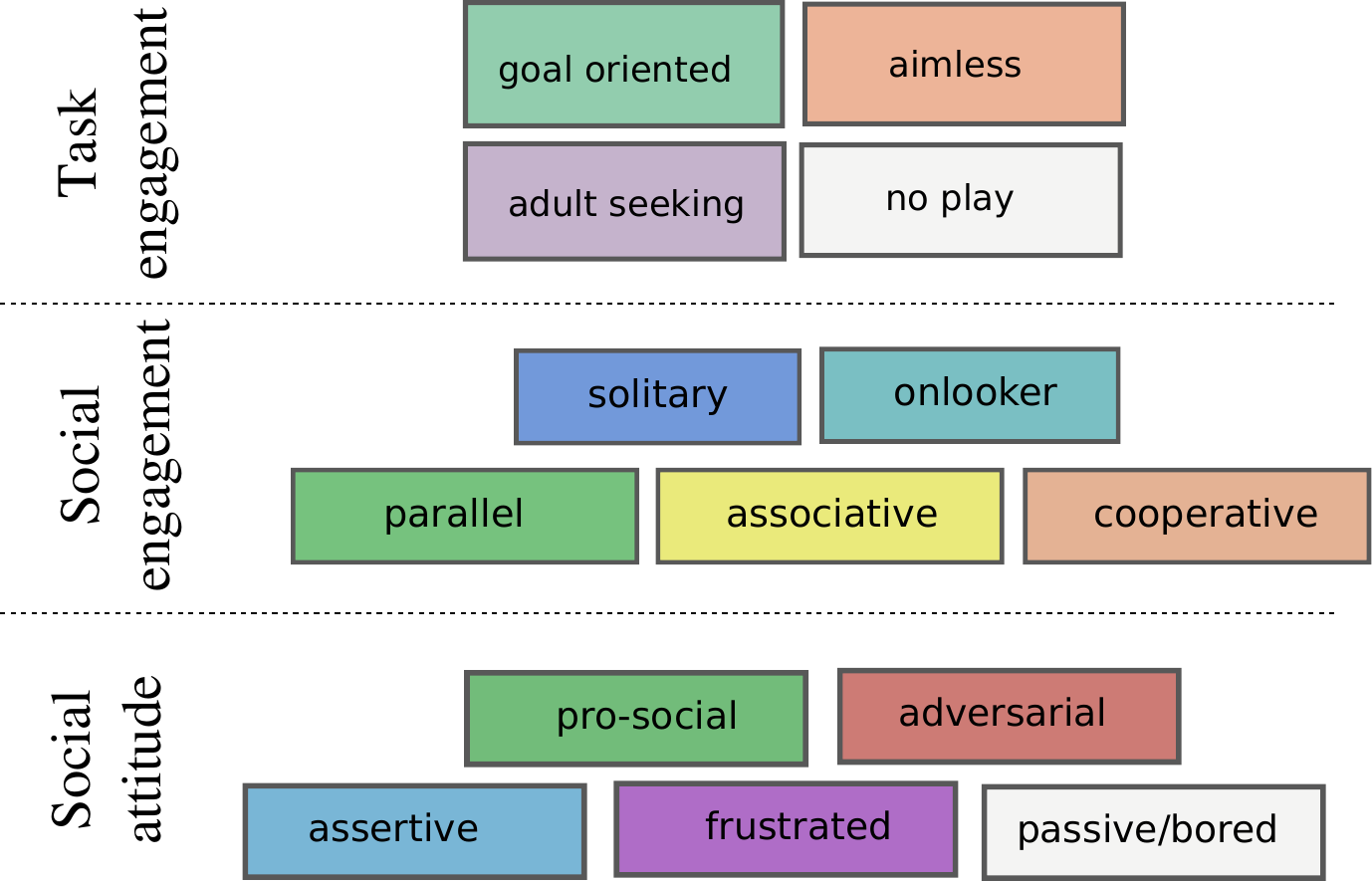}
    \caption{The coding scheme used for annotating social interactions occurring
    during free-play episodes. Three main axis are studied: task engagement,
    social engagement and social attitude.}
    \label{fig|coding-scheme}
\end{figure}

Our aim is to provide insights on the social dynamics, and we have synthesised a
new coding scheme for social interactions that reuse and adapt established
social scales. Our coding scheme (Figure~\ref{fig|coding-scheme}) looks specifically
at three axis: the level of \emph{task engagement} (that distinguishes between
\emph{focused}, \emph{task oriented} behaviours, and \emph{disengaged} -- yet
sometimes highly social -- behaviours); the level of social engagement (reusing
Parten's stages of play, but at the micro-task level); the social attitude (that
encode attitudes like \emph{supportive}, \emph{aggressive}, \emph{dominant},
\emph{annoyed}, etc.)

\paragraph{Task engagement}

The first axis of our coding scheme aims at making a broad distinction between
`on-task' behaviours (even tough the free-play sandbox does not explicitly
require the children to perform a specific task, they are still engaged in an
underlying task: to play with the game) and `off-task' behaviours. We call
`on-task' behaviours \emph{goal oriented}: they encompass considered, planned
actions (that might be social or not). \emph{Aimless} behaviours (with respect
to the task) encompass opposite behaviours: being silly, chatting about
unrelated matters, having a good laugh, etc. These \emph{Aimless} behaviours are
in fact often highly social, and play an important role in establishing trust
and cooperation between the peers. In that sense, they should not be discarded.

\paragraph{Social engagement: Parten's stages of play at micro-level}

In our scheme, we characterise \emph{Social engagement} by building upon
Parten's stages of play. These 5 stages of play are normally
used to characterise rather long sequences (at least several minutes) of social
interactions. Here, we apply them at the level of each of the micro-sequences of
the interactions: one child is drawing and the other is observing is labelled as
\emph{solitary play} for the former child, \emph{on-looker} behaviour for the
later; the two children discuss what to do next: this sequence is annotated as
a \emph{cooperative} behaviour; etc.

By suggesting such a fine-grained coding of social engagement, we enable proper
analyses of the internal dynamics of a long sequence of social interaction.

\paragraph{Social attitude}

The constructs related to the social \emph{attitude} of the children derive from the
\emph{Social Communication Coding System} (SCCS) proposed by
Olswang~\etal\cite{olswang2006reliability}.  The SCCS consists in 6 mutually
exclusive constructs characterising social communication (\emph{hostile};
\emph{pro-social}; \emph{assertive}; \emph{passive}; \emph{adult seeking};
\emph{irrelevant}) and were specifically created to characterise children
communication in a classroom setting.

We transpose these constructs from the communication domain to the general
behavioural domain, keeping the \emph{pro-social}, \emph{hostile} (whose scope
we broaden in \emph{adversarial}), \emph{assertive} (\ie dominant), and
\emph{passive} constructs. In our scheme, the \emph{adult seeking} and
\emph{irrelevant} constructs belong to Task Engagement axis.

Finally, we have added the construct \emph{Frustrated} to describe children who
are reluctant or refuse to engage in a specific phase of interaction
because of a perceived lack of fairness or attention from their peer, or because
they fail at achieving a particular task (like a drawing).

\paragraph{Video coding}

\begin{figure}
    \centering
    \includegraphics[width=\columnwidth]{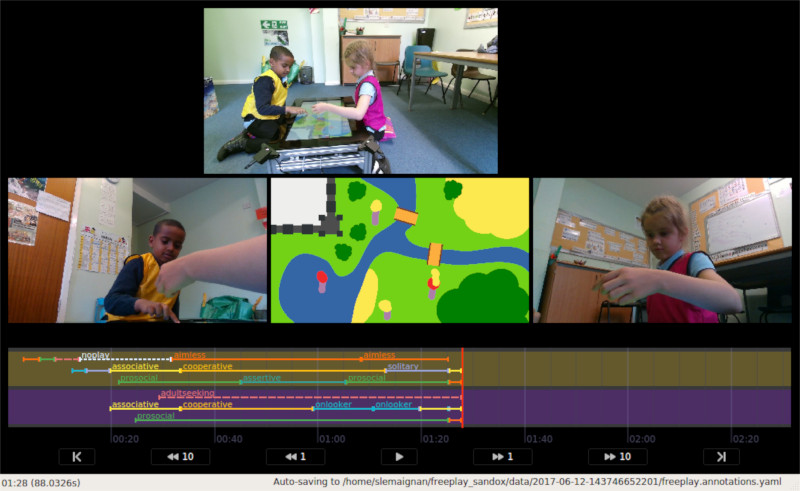}
    \caption{Screenshot of the dedicated tool developed for rapid annotation of
    the social interactions.}
    \label{annotator}
\end{figure}

The coding is performed post-hoc with the help of a dedicated annotation tool
(Fig.~\ref{annotator} which is part of the free-play sandbox toolbox. This tool
can replay and randomly seek in the three video streams, synchronised with the
recorded state of the game (including the drawings as they are created). An
interactive timeline displaying the annotations is also displayed.

The annotation tool offers a remote interface for the annotator (made of large
buttons, and visually similar to Figure~\ref{fig|coding-scheme}) that is
typically displayed on a tablet and allow the simultaneous coding of the
behaviours of the two children.  Usual video coding practices (double-coding of
a portion of the dataset and calculation of an inter-judge agreement score)
would have to be followed.

\section{Baseline Datasets}
\label{sec:dataset}

We have been using the free-play sandbox task for an initial, large scale, data
collection over a period of 3 months during Spring 2017.

This campaign aimed at (1) extensively evaluating the task itself (would
children engage and exhibit a large range of social dynamics and behaviours?),
(2) making sure the whole software architecture and data acquisition pipeline
were reliable (they were), and (3) establishing two experimental baselines for
the free-play sandbox task: the `human' baseline on one hand (child-child
condition), an `asocial' baseline on the other hand (child - \emph{non-social}
robot condition). These two baselines are situated at the two ends of the
spectrum of social interaction. They aim at characterising the qualitative and
quantitative bounds of this social spectrum and can be used by the research
community to evaluate given interaction policies.

A detailed description of the dataset is outside of the scope of this paper, and
we only provide hereafter cursory informations on the dataset. Specific details
regarding the methodology and the acquisition procedure can be found on the
dataset website\footnote{\url{https://freeplay-sandbox.github.io/}}. The dataset is open
and accessible to any interested researcher, subject to adequate ethical clearance.

In total, 120 children were recorded for a total duration of 45 hours and 48
minutes of data collection. These 120 children (age 4 to 8) were split into two
conditions: a child-child condition and a child-robot condition. In both
condition, and after a short tutorial, the children were simply invited to
freely play with the sandbox, for as long as they wished (with a cap at 40 min).

In the child-child condition (as seen in Figure~\ref{annotator}), 45 free-play interactions (\ie 90 children) were
recorded with a duration M=24.15 min (SD=11.25 min).

In the child-robot condition, 30 children were recorded, M=19.18 min (SD=10
min). In this later condition, the robot
behaviour was coded to be purposefully \emph{asocial}: the robot would autonomously play with
the game items, but would avoid any social interaction (no social gaze, no
verbal interaction, no reaction to the child-initiated game actions).

Over the dataset, the children faces are detected on 98\% of the images, which
validates the location of the camera and the children to use the cameras to obtain
facial social features.

\begin{figure}
    \centering
    \includegraphics[width=0.9\columnwidth]{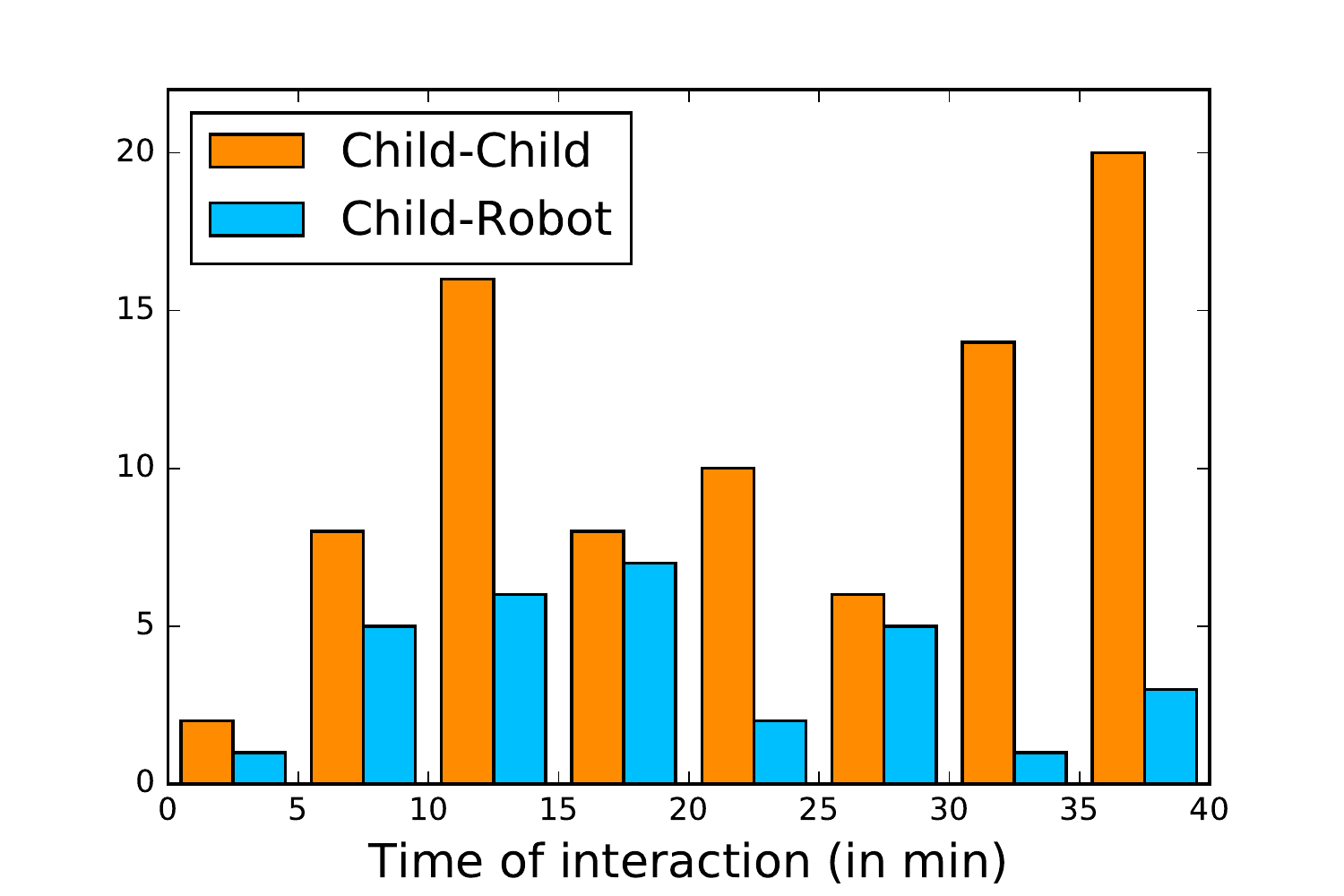}
    \caption{Durations of the interactions for the two conditions.}

    \label{fig|durations}
\end{figure}

Figure~\ref{fig|durations} presents an histogram of the durations of the
interactions for the two baselines. The distribution of the child-child
interaction durations shows that (1) all children
engage easily and for non-trivial amounts of time with the task; (2) the task
leads to a wide range of level of commitment, which is desirable: it supports
the claim that the free-play sandbox is an effective paradigm to observe a range
of different social behaviours; (3) long interactions (>30 min) can result,
which is especially desirable to study social dynamics.

In contrast, and notwithstanding the smaller number of participants, the
distribution of the child-robot interaction durations shows these interactions are
generally shorter. This is expected as the robot was explicitly programmed not
to interact with the children, resulting in a rather boring (and at time,
awkward) situation where the child and the robot where playing side-by-side --
in some case for rather long periods of time -- without interacting at all.


\section{Discussion \& Conclusion}
\label{sec:discussion}

\subsection{Analysis of the free-play sandbox}

The free play sandbox elicits a \emph{loosely structured} form of play: the
actual play situations are not known and might change several times during the
interaction; the game actions, even though based on a single interaction
modality (the touchscreen), are varied and unlimited (especially when
considering the drawings); the social interactions between participants
are multi-modal (speech, body postures, gestures, facial expressions, etc.) and
unconstrained. This loose structure creates a fecund environment for children to
express a range of complex, dynamics, natural social behaviours that are not
tied to an overly constructed social situation.

The interaction is loosely structure. It is nonetheless structured:
First, the physical bounds of the sandbox (an interactive table) limit the
play area to a well defined and relatively small area. As a consequence,
children are mostly static (they are sitting in front of the table) and their
primary form of physical interaction is based on 2D manipulations on a screen.

Second, the game items themselves (visible in Figure~\ref{fig|sandbox})
structure the game scenarios. They are iconic characters (animals or children)
with strong semantics associated to them (like 'crocodiles like water and eat
children'). The game background, with its recognizable zones, also elicit a
particular type of games (like building a zoo or pretending we explore the
savannah).

These elements of structure (along with other, less important, ones) make it
possible for the free-play sandbox paradigm to retain some key properties that
makes it a practical and effective scientific tool: because the game builds on
simple and universal play mechanics (drawings, pretend play with characters),
the paradigm is essentially cross-cultural; because the sandbox is physically
bounded and relatively small, it can be easily transported and practically
deployed in a range of environments (schools, exhibitions, etc.); because the
whole apparatus is well defined and relatively easy to duplicate (it essentially
consists in one single touchscreen computer), the free-play sandbox facilitates
replication of findings in HRI while preserving ecological validity.

\subsection{Towards the machine learning of social interactions?}
\label{sec:conclusion}

We presented a set-up and data set of relatively unconstrained interaction
between children and between a robot and a child. The set-up captures a rich set
of multimodal streams which can be used to mine the social, verbal and
non-verbal communication between two parties engaging in a rich free-play
interaction. The data holds considerable promise for training social signal
interpretation software, such as engagement interpretation or eye gaze reading.
The dataset collected has sufficiently rich data and a wide range of multi-modal
dimensions making it particularly suitable for Deep Learning of social signal
processing algorithms. It also allow for very rich input to action selection
mechanisms needed for autonomous robot behaviour. Future work will focus on
mining the data for social patterns occurring in play situations, as per
Parten's classification, and will attempt to extract social signals relevant to
drive the interaction. Some early results show, for instance,  that deep
learning shows considerable promise for high-resolution tracking of eye gaze
from the RGB video streams.

\section*{Acknowledgments}

This work has been supported by the EU H2020 Marie Sklodowska-Curie Actions
project DoRoThy (grant 657227).

\bibliographystyle{ACM-Reference-Format}
\bibliography{biblio}
\end{document}